\documentclass[runningheads]{llncs}
\usepackage{graphicx}
\usepackage{subcaption}
\usepackage{array}
\usepackage{todonotes}
\usepackage[skip=3mm]{caption}
\usepackage{caption}
\usepackage{url}
\usepackage{hyperref}
\usepackage{subcaption}

\newlength{\twosubht}
\newsavebox{\twosubbox}

\begin{document}
\title{Transfer Learning  and Curriculum Learning\\ in Sokoban}
%
%
\author{
Zhao Yang\inst{1}
\and
Mike Preuss\inst{2}
\and
Aske Plaat\inst{3}
}
\authorrunning{Z. Yang et al.}
%
\institute{LIACS, Leiden University, the Netherlands \\
\email{z.yang@liacs.leidenuniv.nl}
\and
LIACS, Leiden University, the Netherlands \\
\email{m.preuss@liacs.leidenuniv.nl}
\and
LIACS, Leiden University, the Netherlands \\
\email{aske.plaat@gmail.com}}
\maketitle              
\begin{abstract}
Transfer learning can speed up training in machine learning, and is regularly used in classification tasks. It reuses prior knowledge from other tasks to pre-train networks for new tasks.
In reinforcement learning, learning  actions for a behavior policy that can be applied to new environments is still a challenge, especially for tasks that involve much planning. 
Sokoban is a challenging puzzle game. It has been used widely as a benchmark in planning-based reinforcement learning. In this paper, we show how prior knowledge improves learning in Sokoban tasks. We find that reusing feature representations learned previously can accelerate learning new, more complex, instances. In effect, we show how curriculum learning, from simple to complex tasks, works in Sokoban. Furthermore, feature representations learned in simpler instances are more general, and thus lead to positive transfers towards more complex tasks, but not vice versa. We have also studied which part of the knowledge is most important for transfer to succeed, and identify which layers should be used for pre-training.\footnote{We open-sourced all the code we used. It can be found, after the review, at\\ 
\url{https://anonymous.org/blind-review}
}

\end{abstract}
\section{Introduction}

Humans are good at reusing prior knowledge when facing new problems. As a consequence, we learn new tasks quickly, a skill of great interest in machine learning.  In the human brain, information received by our sensors is first transformed into different forms, and different types of transformed information are stored in different areas of our brain. When another problem arrives later on, we retrieve useful information and adjust it to better suit solving this new problem.
The knowledge stored in artifical neural networks is also re-usable and transferable~\cite{yosinski2014transferable}. 
In supervised learning, pre-trained networks are commonly applied in computer vision~\cite{he2016deep,simonyan2014very} and natural language processing~\cite{brown2020language,devlin2018bert}. Feature representations learned from images or words overlap to some extent, which makes such feature representations reusable and transferable. In reinforcement learning (RL), transfer learning is relatively new, although with the spread of deep neural networks, reusing pre-trained models becomes possible in RL as well~\cite{anderson2015faster,cruz2017pre}. Transfer learning works well in RL for recognition tasks, but tasks that rely heavily on planning are harder.

\begin{figure}[!t]
    \centering
    \includegraphics[scale=0.5]{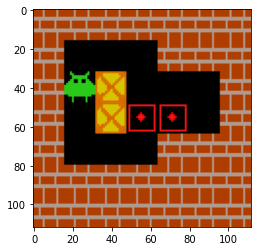}
    \caption{An example instance of Sokoban.}
    \label{fig:example_2}
\end{figure}
In this paper, we study transfer learning of behavior in Sokoban, a popular RL game in which planning is important~\cite{dor1999sokoban,feng2020solving}. It has already been proved that Sokoban is PSPACE-complete~\cite{culberson1997sokoban} and NP-hard problem~\cite{dor1999sokoban}.  An example instance from~\cite{racaniere2017imagination} is shown in Fig.~\ref{fig:example_2}. The goal of Sokoban is to control a warehouse worker that pushes all boxes onto targets. Sokoban is a challenging game where one wrong move can lead to a dead end (after a box has been pushed, it can not be pulled, and we cannot undo an inadvertent push). 
This non-reversibility is known to make games harder for AI agents \cite{CookR19}.
Learning to solve Sokoban tasks is a challenge, especially in the multi-box scenario. For humans, if we have  learned the basics of Sokoban (what is a box, what can an agent do), and if we are faced with a new, more complex instance, then we immediately focus on the new challenges in the instance,  rather than re-learning the basics again. This building on prior knowledge saves time in the problem-solving process.

We investigate
if we can achieve this kind of  pretraining/fine-tuning learning in an RL agent.
Our main hypothesis is  that feature representations learned in Sokoban instances can be reused to improve solving other instances, and that features  learned in simpler instances are more general and better transferable.
We test this hypothesis by means of different experiments, in which parts of the neural network that has previously been trained on one type of instances (e.g. one box one target) are taken over (unchanged) to a new type of instances (e.g. two boxes two targets), whereas the remaining part of the network is trained on these new instances from scratch. The overall idea is that we see successful transfer if the preserved knowledge (in terms of network layers) leads to a faster learning process on the new problem type.

The main contributions of this paper are as follows: First, we show that feature representations learned in simple Sokoban instances can accelerate  learning in more complex instances, indicating that curriculum learning can be used in Sokoban. 
Second, feature representations of simpler instances are more general and reusable than features learned in more complex instances. Third, our results confirm that in RL lower layers learn more general features. Interestingly, in some cases the best performance is achieved when more specific features are transfered, when  source task and target task are similar enough to support 
these more specific features. Fourth, we  found negative transfer from a simple supervised learning task, which tells us that choice and design of the source tasks are crucial. Fifth, we show that transferring top-fully-connected layers will not only be unhelpful but also harmful to the learning.
We also used popular visualization techniques to explore potential reasons for successful transfers, which we explain in detail. Our code and test environments will be made available after blind review. 

The paper is structured as follows: we first briefly review related work on  transfer learning and Sokoban in the next section; 
then the environment and methods we are using are described in Section~\ref{methods}; Section~\ref{experiments} shows the experimental settings and results; in the last section, we conclude our work and discuss some potential future directions.

\section{Related Work}
\label{related}
De la Cruz et al.~\cite{de2016initial} studied the reuse of feature representations between two similar games: Breakout and Pong, using DQN. They used a  3-layer convolutional network. Weights learned in one game were transferred to improve learning the other game; results showed positive transfer of features between the different games. Pong and Breakout do not require planning; in our experiments, in Sokoban, we study how a curriculum  of simpler instances can benefit the learning of complex instances.   Spector et al.~\cite{spector2018sample} used self-transfer in a DQN grid-world task to identify  which parts should be transferred and which parts should be fixed, showing significant benefit of knowledge transfer. 

Sokoban is a planning task that has been used as a benchmark for
model-based reinforcement learning~\cite{racaniere2017imagination,hamrick2020role}. It  It has also been used in model-free RL~\cite{guez2019investigation,guez2018learning}, achieving   performance competitive with  model-based methods. The efficiency of AlphaZero-style curriculum learning has been shown by solving hard single Sokoban instances~\cite{feng2020novel,feng2020solving}. Previous works were aimed at solving single Sokoban instances; our paper focuses on the transferability of learned knowledge among \emph{different} instances.

This transferability of learned feature representations was first studied in image classification problems~\cite{yosinski2014transferable}. It was shown that bottom layers in CNNs extract more general features while ones extracted from back layers are more specific. In this paper, we  verify this idea under RL settings.

Reinforcement learning~\cite{sutton2018introduction,plaat2020learning} aims to reinforce behaviors of the learning agent by rewarding signals obtained from interactions with the environment. It has reached super-human performance in games such as Go~\cite{silver2017mastering}, StarCraft~\cite{vinyals2019grandmaster,ontanon2013survey}, as well as Atari games~\cite{badia2020agent57} and robotic tasks. In this paper we follow the conventional MDP notation for RL~\cite{sutton2018introduction}. 

Transfer learning reuses prior knowledge  to improve the learning efficiency or performance in new tasks~\cite{xu2020transfer,taylor2009transfer}. In reinforcement learning, higher-level knowledge such as macro actions, skills and lower-level knowledge such as reward functions, policies could be transferred. Transferring learned knowledge could take different approaches, such as reward shaping~\cite{brys2015policy}, learning from demonstration~\cite{nair2018overcoming} and policy reuse~\cite{fernandez2010probabilistic}. 

\section{Experimental Setup}
\label{methods}


\begin{figure}[!b]
\begin{subfigure}{.3\textwidth}
  \centering
  \includegraphics[width=.8\linewidth]{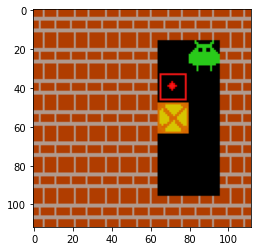}
\end{subfigure}%
\begin{subfigure}{.3\textwidth}
  \centering
  \includegraphics[width=.8\linewidth]{figs/example_2.png}
\end{subfigure}
\begin{subfigure}{.3\textwidth}
  \centering
  \includegraphics[width=.8\linewidth]{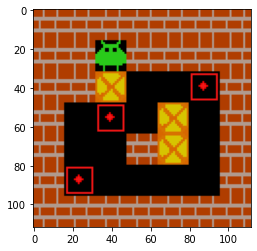}
\end{subfigure}
\caption{Examples of Sokoban instances, increasing in difficulty from 1 box and 1 target to 3 boxes and 3 targets}
\label{fig:sokoban_example}
\end{figure}

The environment used in the paper is the Gym environment for Sokoban~\cite{SchraderSokoban2018}; for the agent algorithms we follow Weber et al.~\cite{racaniere2017imagination}. Examples are shown in Fig.~\ref{fig:sokoban_example}. The game is solved by controlling the agent (green sprite) to push all boxes (yellow squares) onto corresponding targets(red squares). There's no hint about which boxes should on which targets, and boxes can only be pushed;  some actions are irreversible, and can leave the game in an unsolvable state. The difficulty of the game can be increased easily by putting more boxes as well as targets into generated rooms. The agent can  go up, down, left, and right. The agent gets a final reward of 10 by pushing all boxes on targets. Pushing a box on a target will result a reward of 1 and a penalty of -1 for pushing a box off a target. We also give a small penalty of 0.1 for each step the agent takes.


We perform three types of experiments: (1) related tasks (source and target tasks are both RL tasks, while source tasks are to solve $n$-boxes Sokoban instances and target tasks are to solve $m$-boxes Sokoban instances, where $n\neq m$), (2) different tasks (source tasks are SL tasks and target tasks are RL tasks), and (3) different texture appearance(source and target tasks are both RL tasks, while source tasks are to solve original Sokoban instances and target tasks are to solve Sokoban instances with different texture appearance). The agent was first pre-trained on source tasks and then fine-tuned on target tasks. RL tasks are to solve 100 randomly generated $n$-boxes Sokoban instances. SL tasks are to recognize the location of the agent in Sokoban instances. 

\begin{figure}[!ht]
    \centering
    \includegraphics[scale=0.6]{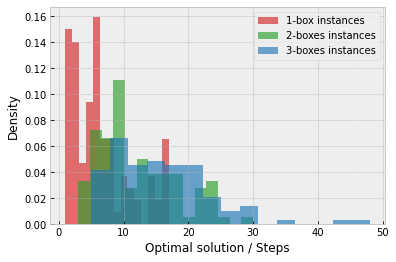}
    \caption{Distribution of optimal solutions in different Sokoban instances.}
    \label{fig:opt_maps}
\end{figure}

The overall statistics of the maps are shown in Fig.~\ref{fig:opt_maps}. As the number of objectives increases, the number of steps for the optimal solution also increases, and so does the difficulty of solving the game.

\subsection{Neural Network Architecture}
The neural network we employ is taken from the DeepMind baseline~\cite{racaniere2017imagination} directly without  hyper-parameter tuning. The model consists of 3 convolutional (Conv) layers with kernel size 8x8, 4x4, 3x3, strides of 4, 2, 1, and number of output channels 32, 64, 64. This is followed by a fully connected (FC) hidden layer with 512 units. The outputs of this FC layer will be fed into two heads: one for outputting the policy logits and one for outputting the state value. This is one of the most commonly-used architectures in RL, we selected it also in order to show what can be achieved with popular architecture. Details of architecture and hyper parameters we employ are found in Table~\ref{tab:hp}.

\begin{table}[!ht]
    \centering
    \begin{tabular}{|l|r|}
        \hline
        learning rate & $7\cdot 10^{-4}$\\
        \hline
        gamma & 0.99\\
        \hline
        entropy coef & 0.1\\
        \hline
        value loss coef & 0.5\\
        \hline
        eps & $10^{-5}$\\
        \hline
        alpha & 0.99\\
        \hline
        rollout storage size & 5\\
        \hline
        No. of environments for collecting trajectories & 30 \\
        \hline
    \end{tabular}
    \caption{Hyper-parameters of the neural network and training.}
    \label{tab:hp}
\end{table}

\subsection{Transfer Approach}
\begin{figure}[!t]
\begin{subfigure}{.31\textwidth}
  \centering
  \includegraphics[width=.95\linewidth]{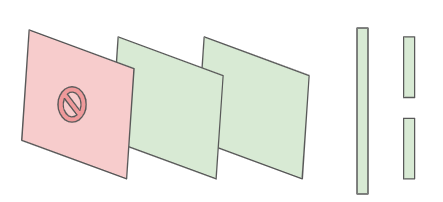}
\end{subfigure}%
\begin{subfigure}{.31\textwidth}
  \centering
  \includegraphics[width=.95\linewidth]{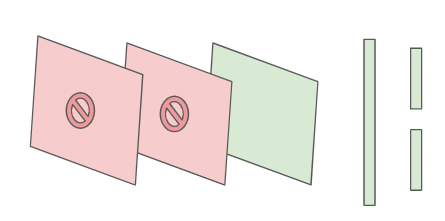}
\end{subfigure}
\begin{subfigure}{.38\textwidth}
  \centering
  \includegraphics[width=.75\linewidth]{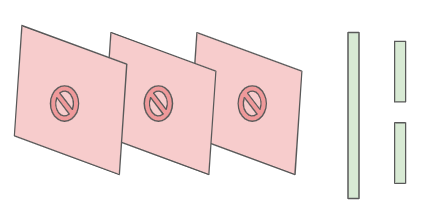}
\end{subfigure}
\caption{Three different transfer approaches, red layers are fixed while green layers are trainable. They correspond $k=1,2,3$ from left to right respectively.}
\label{fig:transfer_vis}
\end{figure}

The main idea of our transfer approach is to reuse feature representations from source tasks learned by the Conv layers  in new unseen target tasks. As detailed in the last sub-section, our model consists of 3 Conv layers and 2 FC layers. The feature representations were transferred to new tasks by copying the weights of the first $k$ Conv layers trained in source tasks (where there are $n_s$ boxes/targets) to initialize the new learning model in target tasks (where there are $n_t$ boxes/targets). Then we froze these weights (they were no longer trainable) and retrained the remaining part of the model. In our experiments, $k \in \{1,2,3\}$, $n_s \in \{1,2,3\}$, $n_t \in \{1,2,3\}$. Please refer to Fig.~\ref{fig:transfer_vis} for an explanation of this approach. Different squares represent different layers of our neural network. The first 3 layers are Conv layers and the last two are FC layers. Reds are weights taken from pre-trained model and fixed, greens are weights reinitialized and trainable.

Solved ratios were used for measuring agents' performances, and they were calculated every 1,000 environment steps. 20 randomly selected test instances were performed by the current learning agent. We say the transfer is \emph{positive} when the performance with the transfer is better than without (training from scratch), and \emph{negative} when the performance with the transfer is worse than without.



\section{Experiments}
\label{experiments}
We designed experiments with different source, target tasks and $k$, in order to verify the hypotheses we proposed. We experimented with Sokoban instances with 1, 2, and 3 boxes. All experiments were run for 1 million environment steps. Experimental details are shown in Table~\ref{tab:exp_detail}. We use abbreviations for each experiment. For instance, \textbf{s}1\textbf{t}1\textbf{k}1 means \textbf{s}ource tasks are \textbf{1}-box instances, \textbf{t}arget tasks are \textbf{1}-box instances and we transfer and fix the \textbf{1}(first) layer. The neural networks were trained using A2C, a single threaded variant of A3C~\cite{mnih2016asynchronous}. All experiments were performed 5 times with different random seeds, and figures were drawn using averaged results with 0.95 confidence interval. Heavy fluctuations were caused by irreversible actions, one irreversible action during the game could make the whole game unsolvable.


\begin{table}[!ht]
    \centering
    \begin{tabular}{|>{\bfseries}c|c|c|c|>{\bfseries}c|c|c|c|}
        \hline
         abbreviation & source task & target task & $k$ & abbreviation & source task & target task & $k$ \\
         \hline
         s1t1k3 & 1-box & 1-box & 3 & s1t2k2 & 1-box & 2-boxes & 2 \\
         \hline
         s2t1k1 & 2-boxes & 1-box & 1 & s1t1k3 & 1-box & 2-boxes & 3 \\
         \hline
         s2t1k2 & 2-boxes & 1-box & 2 & s2t2k3 & 2-boxes & 2-boxes & 3 \\
         \hline
         s2t1k3 & 2-boxes & 1-box & 3 & s1t3k1 & 1-box & 3-boxes & 1 \\
         \hline
         s3t1k1 & 3-boxes & 1-box & 1 & s1t3k2 & 1-box & 3-boxes & 2 \\
         \hline
         s3t1k2 & 3-boxes & 1-box & 2 & s1t3k3 & 1-box & 3-boxes & 3 \\
         \hline
         s3t1k3 & 3-boxes & 1-box & 3 & s2t3k3 & 2-boxes & 3-boxes & 3 \\
         \hline
         sPt1k1 & prediction & 3-boxes & 1 & s1t2k1 & 1-box & 2-boxes & 1\\
         \hline
         s1t1fc\_game2 & 1-box & 1-box\_game2 & fc & & & &\\
         \hline
    \end{tabular}
    \caption{Experimental design details. $k$ is the number of fixed layers.}
    \label{tab:exp_detail}
\end{table}

\subsection{Transfer Among Related Tasks}
\emph{Related tasks} are tasks where  the only difference between source and task is the difficulties of instances, i.e. the number of boxes and targets. (Recall that both source and task are trained on 100 different map-layouts, in all experiments.) 

\begin{figure}[!b]
    \centering
    \includegraphics[scale=0.25]{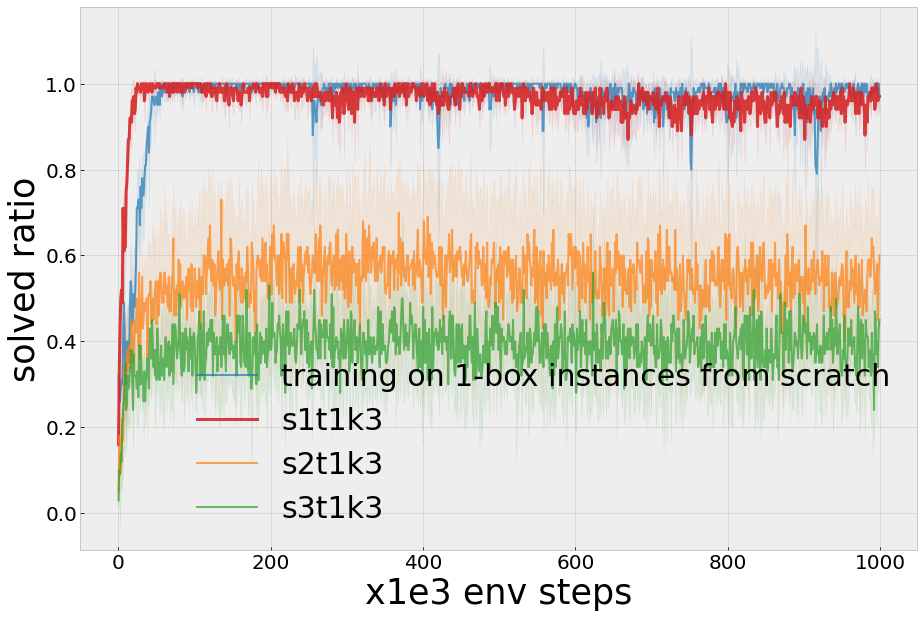}
    \caption{Performance of transferring feature representations learned in 1-box, 2-boxes, 3-boxes instances to learning in 1-box with $k=3$. $n_s=1,2,3$, $n_t=1$, $k=3$. Pre-training on 1-box instances is much better than pre-training on 2 or 3 box instances when training new 1-box instances.}
    \label{fig:training1}
\end{figure}

\begin{figure}[!ht]
\begin{subfigure}{.5\textwidth}
  \centering
  \includegraphics[width=.99\linewidth]{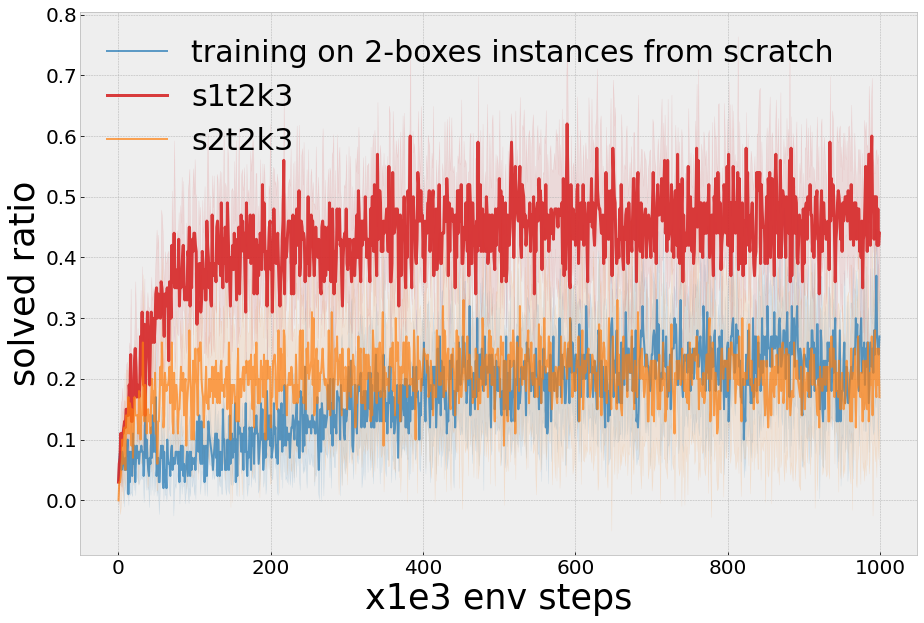}
\end{subfigure}
\begin{subfigure}{.5\textwidth}
  \centering
  \includegraphics[width=.99\linewidth]{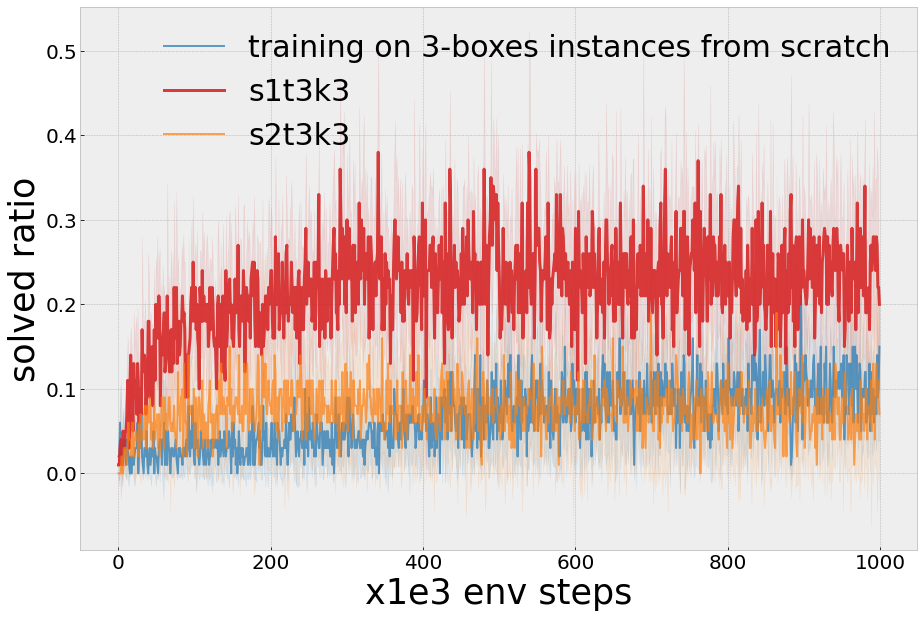}
\end{subfigure}
\caption{Performance of transferring feature representations learned in 1-box, 2-boxes, 3-boxes instances to learning in 2-boxes (left) and 3-boxes (right) with $k=3$. $n_s=1,2,3$, $n_t=2,3$, $k=3$.}
\label{fig:trainings}
\end{figure}

Fig.~\ref{fig:training1} and Fig.~\ref{fig:trainings} show results for training on 1-box, 2-boxes, 3-boxes instances with reusing features learned in different tasks, and we fix $k=3$. All results showed that transferring feature representations learned in single-box instances is positive. Performance of agents (s1t1k3, s1t2k3, s1t3k3) who are using features learned from single-box instances always outperform  other agents, including agents training from scratch and using features learned from  other instances. 
The transfer, however, is not 'bi-directional', feature representations learned in multiple-box instances could not be successfully transferred to the learning in single-box instances. Their performance (s2t1k3, s3t1k3) converged to a relatively low solved ratio, which indicates that transferred features are not suitable for single-box instances. Just as humans learn more general knowledge in simpler cases, our agents also showed that the knowledge learned from single-box instances is more general and transferable than ones learned in multiple-box instances.

\begin{figure}[!t]
\begin{subfigure}{.5\textwidth}
  \centering
  \includegraphics[width=.99\linewidth]{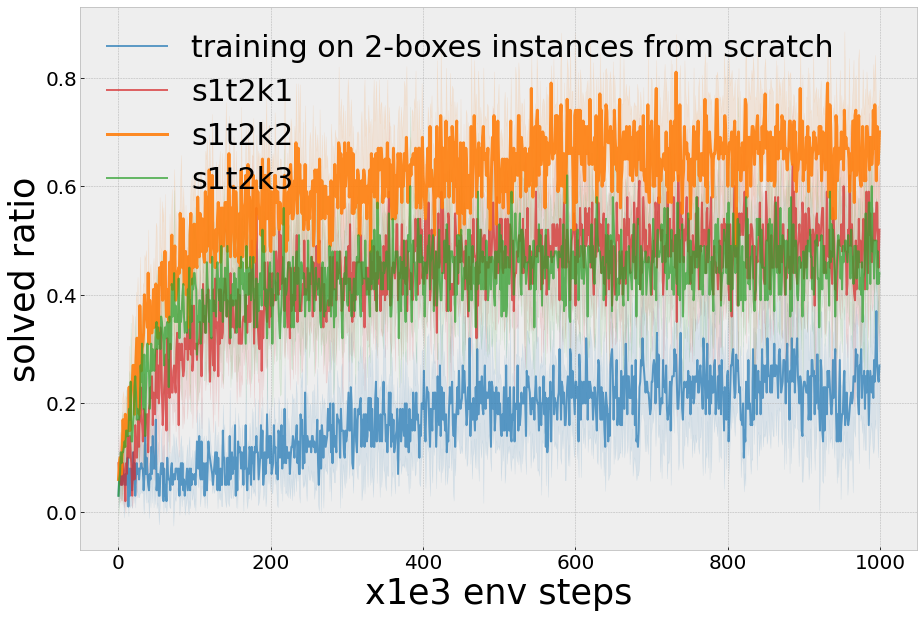}
\end{subfigure}%
\begin{subfigure}{.5\textwidth}
  \centering
  \includegraphics[width=.99\linewidth]{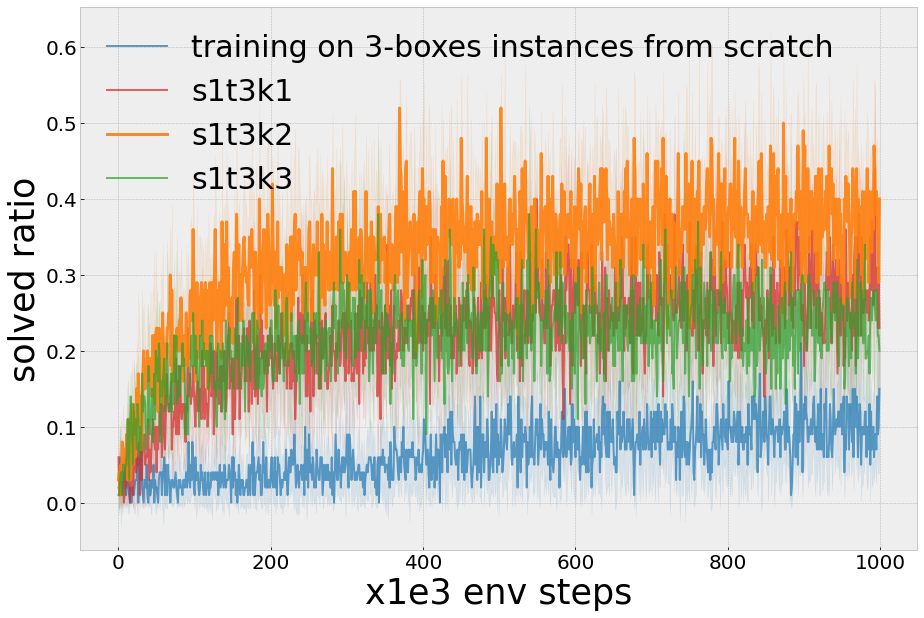}
\end{subfigure}
\caption{Performance of transferring feature representations learned in 1-box instances to learning in 2-boxes (left) and 3-boxes (right) with different $k$. $n_s=1$, $n_t=2,3$, $k=1,2,3$.}
\label{fig:1onN}
\end{figure}

To further enhance performances of transferring features learned in single-box instances, we tried different $k$. We expected that the performance will be the best when $k=1$ since the first layer learn the most general features. However,  the results  in Fig.~\ref{fig:1onN} instead show that not $k=1$ but $k=2$ (s1t2k2, s1t3k2) perform the best. Similar to \cite{de2016initial}, features learned  in the first 2 layers are still general enough for transfer; in addition, the difference between source tasks and target tasks is not as large as expected, and features learned between different instances are more overlapping than  expected.

\begin{figure}[!ht]
\begin{subfigure}{.5\textwidth}
  \centering
  \includegraphics[width=.99\linewidth]{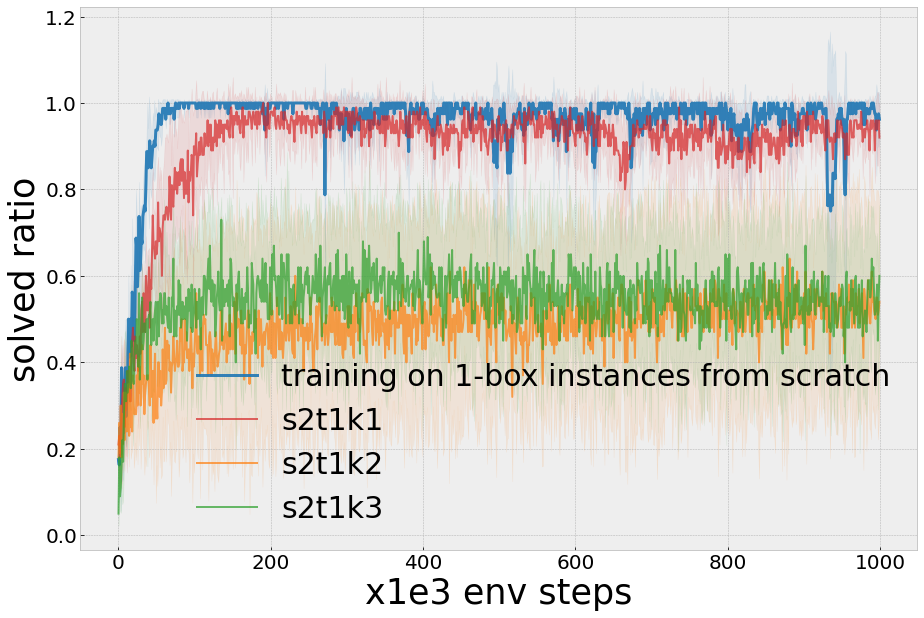}
\end{subfigure}%
\begin{subfigure}{.5\textwidth}
  \centering
  \includegraphics[width=.99\linewidth]{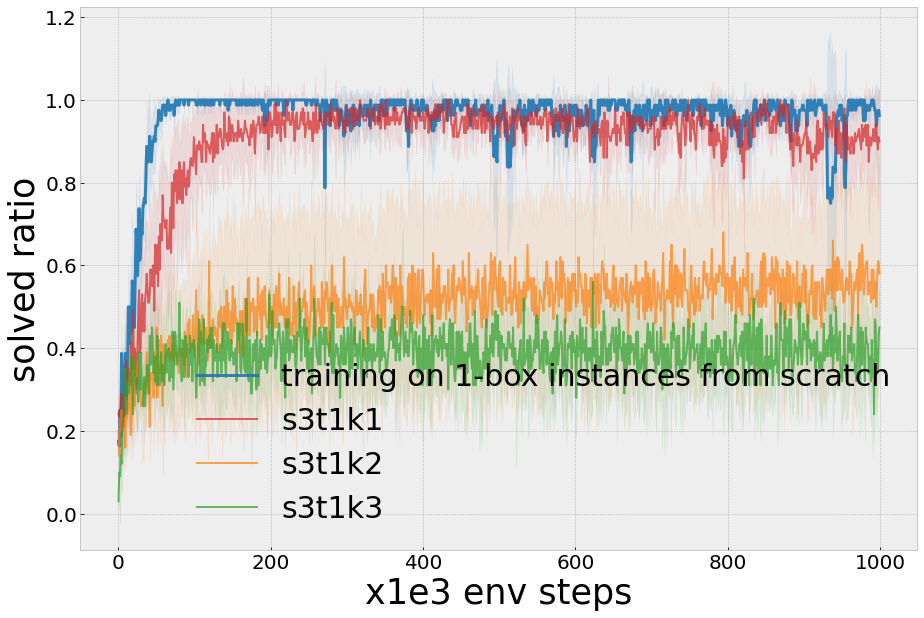}
\end{subfigure}
\caption{Performance of transferring feature representations learned in 2-boxes  (left) and 3-boxes  (right) instances to learning in 1-box instances with different $k$. $n_s=2,3$, $n_t=1$, $k=1,2,3$.}
\label{fig:Non1}
\end{figure}

It is also interesting  to see the influence of how many layers are fixed on the success of the transfer. In particular, we want to know whether a smaller $k$ could change the negative transfer from multiple-box instances to single-box instances into positive. (We believe features from multiple and single-box instances are overlapping to some extent.) Results are shown in Fig.~\ref{fig:Non1}. We see that indeed the first layer (s2t1k1, s3t1k1) did learn enough general features from multiple-boxes instances to solve the single-box instances. Although agents with features only learned by the first layer could converge to decent performance in the end, the transfer is still negative. An interesting point is that $k=3$ (s2t1k3) performs better than $k=2$ (s2t1k2) when source tasks are 2-boxes instances. Note that $k=2$ (s3t1k2) performs better than $k=3$ (s3t1k3) when source tasks are 2-boxes instances. There are more overlapping features between the 2-boxes instances and single instances.

\subsection{Transfer Among Different Tasks (SL/RL)} 

Feature representations learned from previous tasks can either be helpful or harmful. In the previous subsection we saw some positive transfer to related Sokoban tasks, in this subsection we study if transfer between supervised and reinforcement learning tasks works. We follow  prior work, Anderson et al.~~\cite{anderson2015faster} showed that features can be transfered from hand-crafted supervised learning tasks to reinforcement learning. Their model was first trained to predict state dynamics of the environment, and then pre-trained hidden layers were helpful to accelerate solving RL tasks.

\begin{figure}[!ht]
\sbox\twosubbox{%
  \resizebox{\dimexpr.9\textwidth-1em}{!}{%
    \includegraphics[height=3cm]{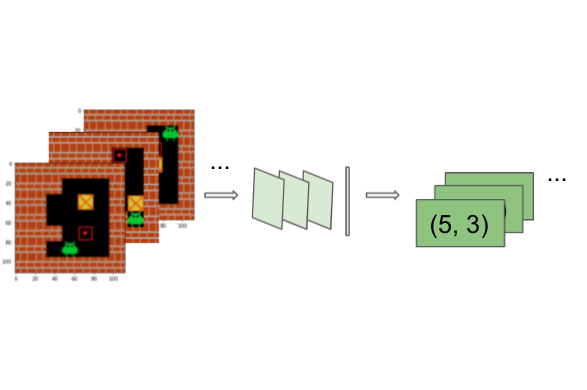}%
    \includegraphics[height=3cm]{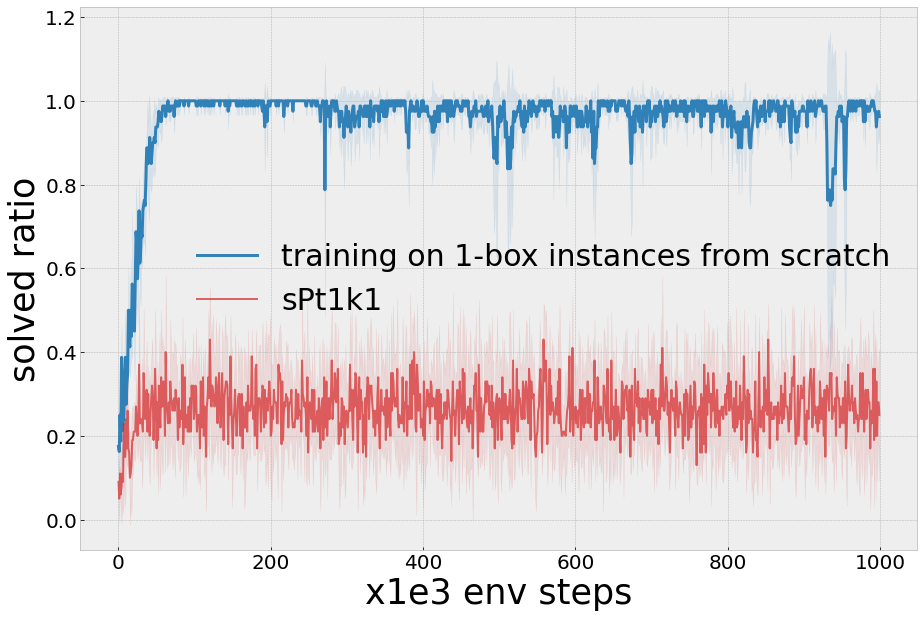}%
  }%
}
\setlength{\twosubht}{\ht\twosubbox}
\centering
\subcaptionbox{\label{fig:prediction}}{%
  \includegraphics[height=\twosubht]{figs/prediction.png}%
}\quad
\subcaptionbox{\label{fig:pre_on_1}}{%
  \includegraphics[height=\twosubht]{figs/Pon1.png}%
}
\caption{(a): How SL tasks work. Input states and neural network will learn to predict locations of the agent. (b): Performance of training from scratch and training with transferred feature representations from SL tasks.}
\end{figure}

For transfer to different (randomly chosen)  instances in Sokoban, we also formed a supervised task, which was to train a prediction model to recognize the location of the agent, shown in Fig.~\ref{fig:prediction}. When humans are solving Sokoban, we first need to know where the agent is before we draw up a plan. If we already know the location of objectives, the solving process could be faster. After the prediction model could correctly recognize where the agent is, we took feature representations of the trained model and plug them into a new agent. The first layer of learned features is fixed, and we only train the remaining part. Fig.~\ref{fig:pre_on_1} shows the performance of transferring and training from scratch. We find  negative transfer for (sPt1k1): the performance is much worse compare with training from scratch. The bad performance is due to the choice of the source task, which is too different for the target tasks, and the expressive power that is learned is inadequate to overcome this difference. 

\subsection{Transfer To Different Appearance}

\begin{figure}[!ht]
\sbox\twosubbox{%
  \resizebox{\dimexpr.9\textwidth-1em}{!}{%
    \includegraphics[height=3cm]{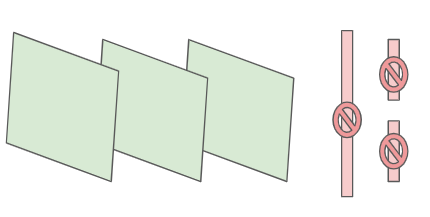}%
    \includegraphics[height=3cm]{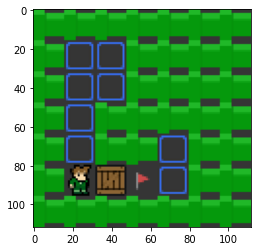}%
  }%
}
\setlength{\twosubht}{\ht\twosubbox}
\centering
\subcaptionbox{\label{fig:fix_fc}}{%
  \includegraphics[height=\twosubht]{figs/fix_fc.png}%
}\quad
\subcaptionbox{\label{fig:game_2}}{%
  \includegraphics[height=\twosubht]{figs/game_2.png}%
}
\caption{(a): Transfer approach for transfer to Game2. FC layers are taken from previously training and fixed, only conv layers will be retrained. (b): An example instance in Game2. We changed appearances in Game2 with different textures of objectives.}
\end{figure}

Experiments we described in previous subsections were all trying to transfer Conv layers which learned feature representations. In the next experiment, we try to make the agent utilize another part of the learned model, which are back FC layers of the whole model. The source and target tasks were both single-box instances, but the target tasks were instances with different appearances. Fig.~\ref{fig:game_2} is an example. The maps used for two groups of tasks were the same, the only difference was how they look like, the appearance was changed, with different textures, and we call it Game2. Fig.~\ref{fig:fix_fc} shows the transfer approach. We took FC layers trained in source tasks and fixed them, and retrained the remaining Conv layers. Since maps were the same, solutions of the instances were the same. When Conv layers learn new feature representations successfully, instances are solved then. 

Fig.~\ref{fig:game2} shows the performance. One would expect that transferred FC layers(s1t1fc\_game2) are  faster because the agent only needs to learn new feature representations. However, the experiments did not show this result. Apparently, when the whole model is trained jointly, it has more flexibility to be trained into the final shape; when the last part of the model is fixed, the learning of the first part will be  trying to cater for the last part in order to solve the problem, which made the learning slower. 

\subsection{Visualizing Agent Detection}
\begin{figure}[!ht]
\sbox\twosubbox{%
  \resizebox{\dimexpr.9\textwidth-1em}{!}{%
    \includegraphics[height=3cm]{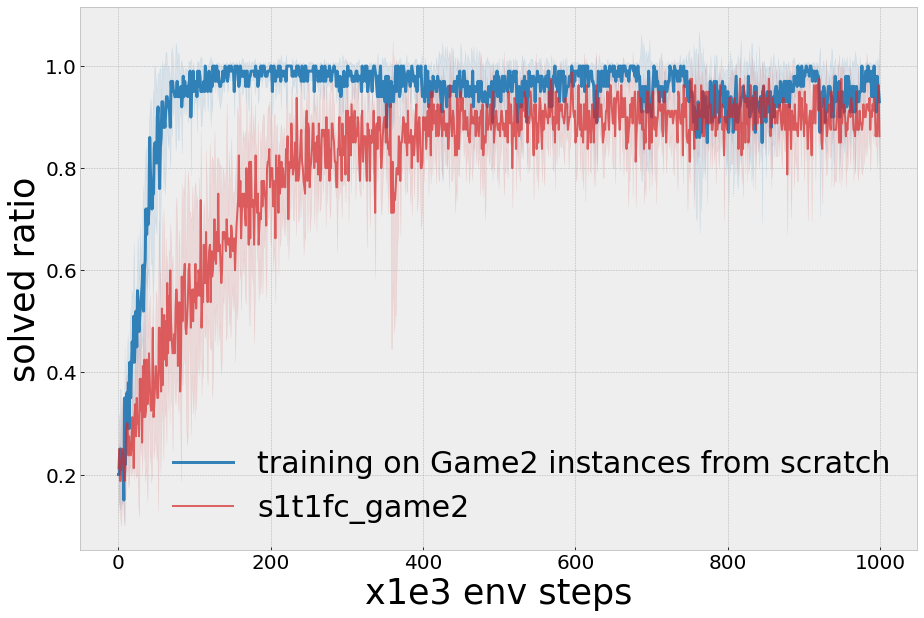}%
    \includegraphics[height=3cm]{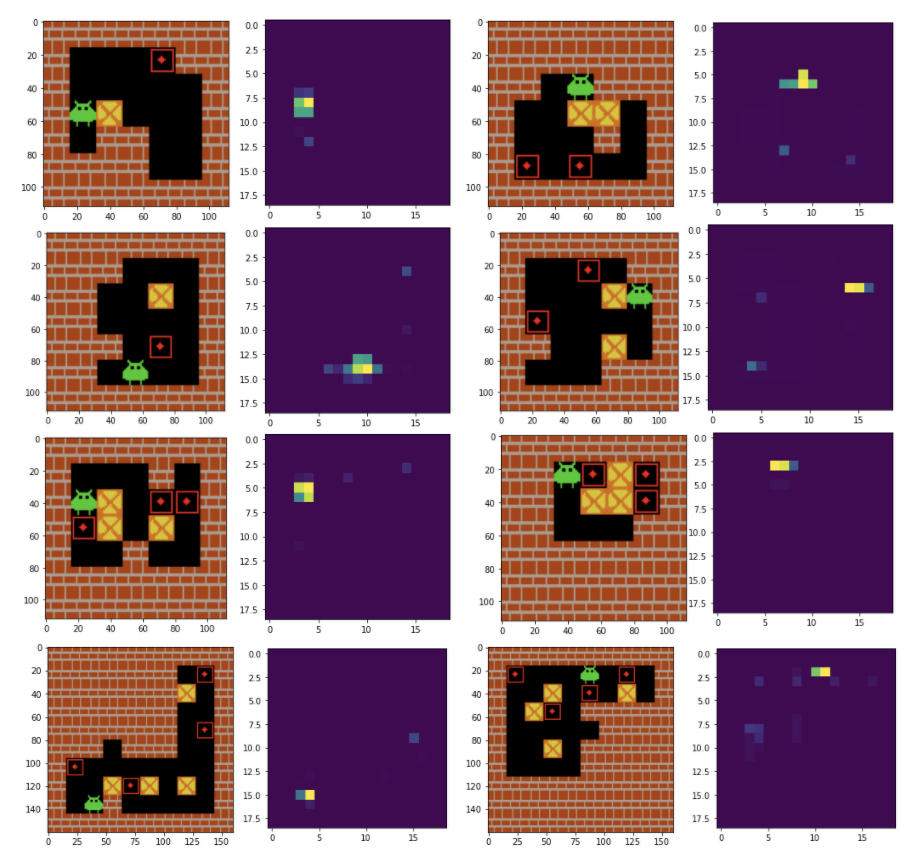}%
  }%
}
\setlength{\twosubht}{\ht\twosubbox}
\centering
\subcaptionbox{\label{fig:game2}}{%
  \includegraphics[height=\twosubht]{figs/game2.png}%
}\quad
\subcaptionbox{\label{fig:agent_detector}}{%
  \includegraphics[height=\twosubht]{figs/agent_detector.png}%
}
\caption{(a): Training on Game2 using transferred FC layers. Its performance is worse than training from scratch. (b): The agent detector. Outputs of the twenty third feature map of the first convolutional layer, which is an agent detector learned from 1-box instances, and it's still usable in multiple-boxes scenarios.}
\end{figure}

In order to better understand what the network learned, we provide a visualization. We follow Yosinski et al.\ who showed that convolutional neural networks can detect latent objectives without explicit labels~\cite{yosinski2014transferable}. Fig.~\ref{fig:agent_detector} shows a latent 'agent detector' for Sokoban. The neural network automatically learned to detect the agent without giving any labels or information. Left rows are pixel inputs, right rows are outputs of one specific feature map. Yellow-green units are detected agents. We note that although the network was trained in single-box instances, it still performed quite well in multiple-box instances, which is a potential reason for the successful transfer. The agent's abilities that were learned in source tasks are useful in target tasks.

\section{Conclusion and Future Work}

Our experiments showed that in a reinforcement learning setting the agent in Sokoban can learn  four characteristics that are similar to humans. (1) Feature representations learned previously can accelerate the new learning in other Sokoban instances. Knowledge learned in previous related tasks could be reused to accelerate new learning, transfer learning is occurring, creating an implicit learning curriculum. (2) Feature representations learned in single-box instances are more general, and are more effective for learning in multiple-boxes instances, but not vice versa. Knowledge learned in simpler tasks is more general and more effective, even in more complex tasks. Further experiments showed negative learning, that confirms these results. (3) Feature representations learned in unrelated supervised learning tasks can  hurt fine-tuning  performance. If the learned knowledge is required to be helpful in new coming tasks, it's better to learn from similar tasks, otherwise the choice of tasks needs to be careful. (4) Fixing the top-fully-connected layers and  retraining the bottom convolutional layers slows down learning and hurts performance. We conclude that learning should have explicit order, less flexibility will not only be unhelpful but also hurt the learning process and the performance.

Our experiments showed that with a simple 5-layer convolutions/fully connected network (based on DeepMind's baseline~\cite{racaniere2017imagination}), transfer learning and curriculum learning of behavior to occur in Sokoban. This is surprising, since Sokoban is a planning-heavy problem,  for which one would expect more elaborate network architectures to be necessary. Reusing pre-trained feature representations in RL fields is not well studied, and to the best of our knowledge, these are the first results show transfer learning and curriculum learning with such a simple network in such a planning-heavy behavioral task. In the future, we would like to see more utilization of pre-trained feature representations and of the enire pre-trained model in RL. We believe that reusing pre-trained model can significantly improve data-efficient reinforcement learning.

\section*{Acknowledgement}
The financial support to Zhao Yang is from the China Scholarship Council(CSC). Computation support is from ALICE and DSLab. The authors thank Hui Wang, Matthias M\"uller-Brockhausen, Michiel van der Meer, Thomas Moerland and all members from the Leiden Reinforcement Learning Group for helpful discussions.

\bibliographystyle{splncs04}
\bibliography{main}

%
%
%

%
\end{document}